\documentclass[conference]{IEEEtran}
\IEEEoverridecommandlockouts
\usepackage{amsmath,amssymb,amsfonts}
\usepackage{algorithmic}
\usepackage[dvipdfmx]{graphicx}
\usepackage[dvipdfmx]{color}
\usepackage{textcomp}
\usepackage{xcolor}
\usepackage{color}
\usepackage{multirow}
\usepackage{wrapfig}
\usepackage{amssymb}
\usepackage{comment}
\usepackage{autobreak}
\usepackage{cite}
\usepackage[normalem]{ulem}

\def\BibTeX{{\rm B\kern-.05em{\sc i\kern-.025em b}\kern-.08em
    T\kern-.1667em\lower.7ex\hbox{E}\kern-.125emX}}
\begin{document}

\title{A New Autoregressive Neural Network Model with Command Compensation for Imitation Learning Based on Bilateral Control
\\
}

\author{\IEEEauthorblockN{1\textsuperscript{st} Kazuki Hayashi}
\IEEEauthorblockA{\textit{Department of Intelligent Interactive Systems} \\
\textit{University of Tsukuba}\\
Tsukuba, Japan \\
s2020780@s.tsukuba.ac.jp}
\and
\IEEEauthorblockN{2\textsuperscript{nd} Ayumu Sasagawa}
\IEEEauthorblockA{\textit{Graduate School of Science and Engineering} \\
\textit{Saitama University}\\
Saitama, Japan \\
a.sasagawa.997@ms.saitama-u.ac.jp}
\and
\IEEEauthorblockN{3\textsuperscript{rd} Sho Sakaino}
\IEEEauthorblockA{\textit{Department of Intelligent Interactive Systems} \\
\textit{University of Tsukuba}\\
Tsukuba, Japan \\
sakaino@iit.tsukuba.ac.jp}
\and
\IEEEauthorblockN{4\textsuperscript{th} Toshiaki Tsuji}
\IEEEauthorblockA{\textit{Graduate School of Science and Engineering,} \\
\textit{Saitama University}\\
Saitama, Japan \\
tsuji@ees.saitama-u.ac.jp}
}

\maketitle

\begin{abstract} 
In the near future, robots are expected to work with humans or operate alone and may replace human workers in various fields such as homes and factories.
In a previous study, we proposed bilateral control-based imitation learning that enables robots to utilize force information and operate almost simultaneously with an expert's demonstration. In addition, we recently proposed an autoregressive neural network model (SM2SM) for bilateral control-based imitation learning to obtain long-term inferences. In the SM2SM model, both master and slave states must be input, but the master states are obtained from the previous outputs of the SM2SM model, resulting in destabilized estimation under large environmental variations. Hence, a new autoregressive neural network model (S2SM) is proposed in this study. This model requires only the slave state as input and its outputs are the next slave and master states, thereby improving the task success rates.
In addition, a new feedback controller that utilizes the error between the responses and estimates of the slave is proposed, which shows better reproducibility.
\end{abstract}

\section{Introduction}
Owing to the aging populations in several countries, an increasing number of robots are expected to be utilized for work instead of humans. To realize robot automation, they are required to perform tasks in diverse environment and while treating various objects. Moreover, motion control is difficult but its demand is rising \cite{b2,b3}. Therefore, machine learning is being used to study motion planning of robots to generate various motions and facilitate easier programming for motion planning. In recent years, two learning methods for robotic manipulation have been primarily researched: reinforcement learning and imitation learning.

In reinforcement learning, robots learn motions autonomously by maximizing the cumulative reward while repeating trials. Although Levine {\it et al.} achieved grasping tasks with reinforcement learning \cite{levine}, the method required excessive trials. A few approaches to solve this issue have been proposed. For example, sim2real is a method of repeating trials in simulated environments instead of real environments \cite{sim2real}. Although this method can reduce the training time, it cannot bridge the learning gap between simulated and real environments, especially when performing tasks that include contact with other objects. Offline reinforcement learning, which utilizes previously collected data without using additional online data, has also been studied to obtain a policy for generating motion without trials in real environments. However, it still has unsolved problems, such as distribution shifts \cite{offline}.

On the other hand, imitation learning is one of the methods that can solve the problems encountered in reinforcement learning and sim2real.
In imitation learning, robots can imitate expert behaviors using direct teaching \cite{survey,survey2}, motion captures \cite{motioncap}, and teleoperation \cite{vr,vr2}. In imitation learning, because robots learn from training data collected during human demonstration, they can learn behaviors more effectively than in reinforcement learning. In recent years, studies that considered both position and force information have been reported and proved to be effective for tasks manipulating deformable objects \cite{f1,f2,f3}.

We propose bilateral control-based imitation learning as a method of using force information \cite{b8,b9,b10}. Bilateral control is a teleoperation method where a human operates the master robot and the slave robot performs tasks in the workspace \cite{bilate1,bilate2,bilate3,b1}. During this period, position synchronization and presentation of the reaction forces are simultaneously executed. Thus, the law of action and reaction is established between the torques of the master and slave. In addition, by introducing bilateral control to imitation learning, the master measures the action force, and the slave measures the reaction force during the data collection phase.
We used bilateral control for data collection, and the method showed effectiveness in tasks that required force adjustment and demonstrated fast motion \cite{b8,b9,b10}. In bilateral control-based imitation learning, an S2M model that predicts the next master state from the current slave state was used.
Another method, autoregressive learning, repeatedly uses predicted values of the model as the next input, and is known to be effective for long-term prediction tasks. Sasagawa {\it et al.} proposed the SM2SM model to adapt autoregressive learning to our bilateral control-based imitation learning \cite{auto}. Consequently, they demonstrated the effectiveness of long-term prediction and a high adaptivity to environmental changes. Although the SM2SM model requires two inputs, the response values of the slave and master, only those of the slave can be measured during the autonomous operation because it operates alone during the autonomous operation. Therefore, the predicted master's state in the previous step is used as the input of the model. This state is the virtual master state, not the actual one, and the system tends to be unstable in large environmental perturbations.

\begin{figure}[t]
\centering
\includegraphics[width=8.0cm]{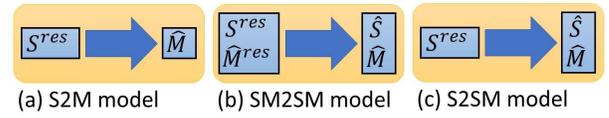}
\caption{Comparison of models}
\label{fig:resmodel}
\end{figure}

\begin{figure}[t]
\centering
\includegraphics[width=7.5cm]{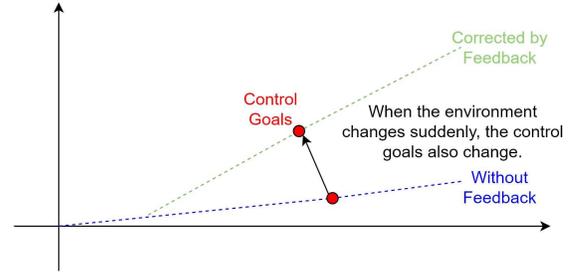}
\caption{Concept of the proposed feedback method}
\label{fig:feed}
\end{figure}

Thus, an S2SM model was proposed in this study. In the S2SM model, the model was trained to predict the next state of the slave and master only from the current slave state. Because the response values of the master are not required in the S2SM model, stable motion generation can be achieved even in sudden environmental changes. In addition, autoregressive learning can be implemented in the S2SM model; therefore, the advantages of the SM2SM model shown by Sasagawa {\it et al.} are not lost. 
As shown in Fig. \ref{fig:resmodel}, the output of the SM2SM and the S2SM model consisted of $\hat S$ and $\hat M$. $S^{res}$ is the response value of the slave, $M^{res}$ is the response value of the master, and ${\hat M}^{res}$ is its estimate. $\hat S$ and $\hat M$ denote the predicted values of the slave state and the master state, respectively. $\hat M$ is also used as the command value of the slave during the autonomous operation. During the autonomous operation in the conventional method, the slave may not be able to achieve desirable behaviors because the perturbation of the environment changes the desired control goals, as shown in Fig. \ref{fig:feed}.
Therefore, a method to adapt to the environmental disturbances is also proposed. Notably, in bilateral control-based imitation learning, the master and slave estimates are strongly related because they are naturally controlled for synchronization. Therefore, in this study, we assumed that the slave’s estimation error is equal to the master’s, and the slave's estimated error is fed back to the master's estimates.

To sum up the above-described issues, two methods are proposed in this paper as follows:
\begin{enumerate}
    \item The S2SM model is used to stabilize the motion generation against sudden environmental changes during the bilateral control-based imitation learning.
    \item A method to is proposed to feed back the estimation errors to suppress them against environmental disturbances.
\end{enumerate}

To validate the effect of the proposed methods, the task of drawing arcs was executed with each model: S2M, SM2SM, and S2SM. The S2SM model was more effective in imitating the experts' behavior because it improved the stability. Furthermore, the force adjustment was realized more precisely due to the addition of the proposed feedback method.

The remainder of this paper is structured as follows. In Section II, the control system and bilateral control are explained. Section III describes the proposed method and imitation learning. Detailed experimental procedures and their results are described in Section IV. Finally, conclusions of the study are presented in Section V.

\section{CONTROL SYSTEM}
The details of the control system are described in \cite{auto}. In addition, we used the same gains and parameters as described in \cite{auto}.

\begin{figure}[t]
\centering
\includegraphics[width=6.0cm]{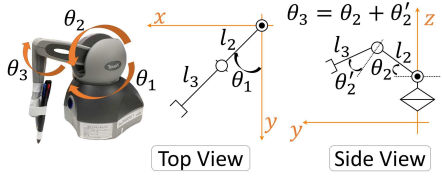}
\caption{Touch$^{TM}$ USB haptic device}
\label{fig:phantom}
\end{figure}

\begin{figure}[t]
\centering
\includegraphics[width=8.0cm]{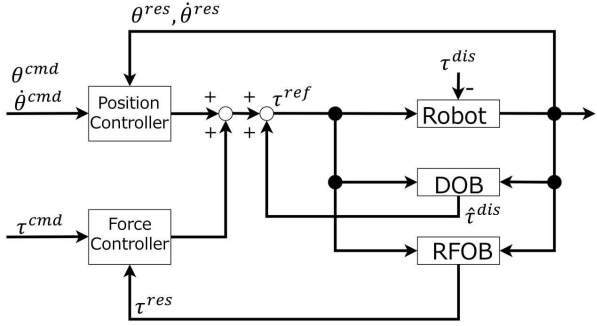}
\caption{Controller}
\label{fig:controller}
\end{figure}

We used two Touch$^{TM}$ USB haptic devices, manufactured by 3D Systems, as shown in Fig.~\ref{fig:phantom}. The joint angles of the manipulator are shown in the center and right side of Fig.~\ref{fig:phantom}. 
The block diagram of the controller is shown in Fig.~\ref{fig:controller}. 
Here, $\theta, \dot\theta$, and $\tau$ represent the joint angle, angular velocity, and robot torque, respectively. The superscripts {\it res}, {\it ref}, and {\it cmd} indicate the response, reference, and command values, respectively. The controller comprised position and force controllers. The position controller consisted of a proportional and a derivative controller, while the force controller included a proportional controller. The manipulators measured the angle $\theta^{res}$ of each joint, $\dot\theta^{res}$ was calculated using a pseudo-derivative, and the disturbance torque $\tau^{dis}$ was estimated with a disturbance observer (DOB)\cite{dob}. Moreover, the torque responses $\tau^{res}$ were calculated using a reaction force observer (RFOB) \cite{rfob}. In this study, robots were operated in a 1 ms control cycle. 

\begin{figure}[t]
\centering
\includegraphics[width=8.0cm]{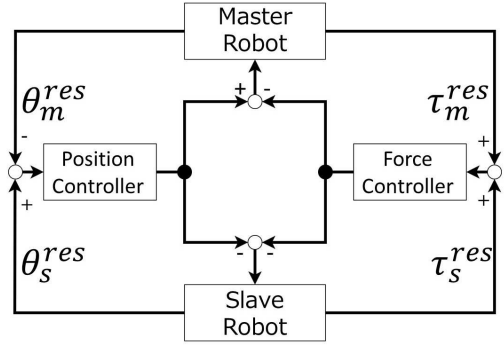}
\caption{4ch bilateral controller used for data collection}
\label{fig:4ch}
\end{figure}

In this study, a 4ch bilateral control was used to collect the training dataset. 4ch bilateral control is a remote control system using two robots: a master and a slave \cite{b8}.
The operator manipulating the master robot can feel the tactile sense generated by the slave robot’s touch. This is because of the synchronized positions of the two robots and the feed back of each other’s forces. The control goals of the 4ch bilateral control are shown in (1) and (2).
In addition, the block diagram satisfying them is shown in Fig.~\ref{fig:4ch},

\begin{eqnarray}
\theta_m^{res} - \theta_s^{res} = 0 , \label{eq:bi1}\\
\tau_m^{res} + \tau_s^{res} = 0 ,\label{eq:bi2}
\end{eqnarray}
where the superscripts $m$ and $s$ indicate the master and slave robots, respectively. These equations hold at all angles and torques. 

\section{BILATERAL CONTROL-BASED IMITATION LEARNING}
 In this section, our approach is explained. In addition, two proposed methods, the S2SM model and the command feedback in autoregressive models, are explained. Imitation learning was executed in the following phases: 
 \begin{enumerate}
  \item Data collection phase,
  \item Training phase,
  \item Autonomous task execution phase.
\end{enumerate}
The details of each phase are described below.

\subsection{Data Collection Phase}
 The training dataset was collected by using a 4ch bilateral control every 1 ms. A human operated the master robot directly, and the teleoperated slave robot performed tasks in the workspace.

\subsection{Training Phase}

\begin{figure}[t]
\begin{center}
\includegraphics[width=9.0cm]{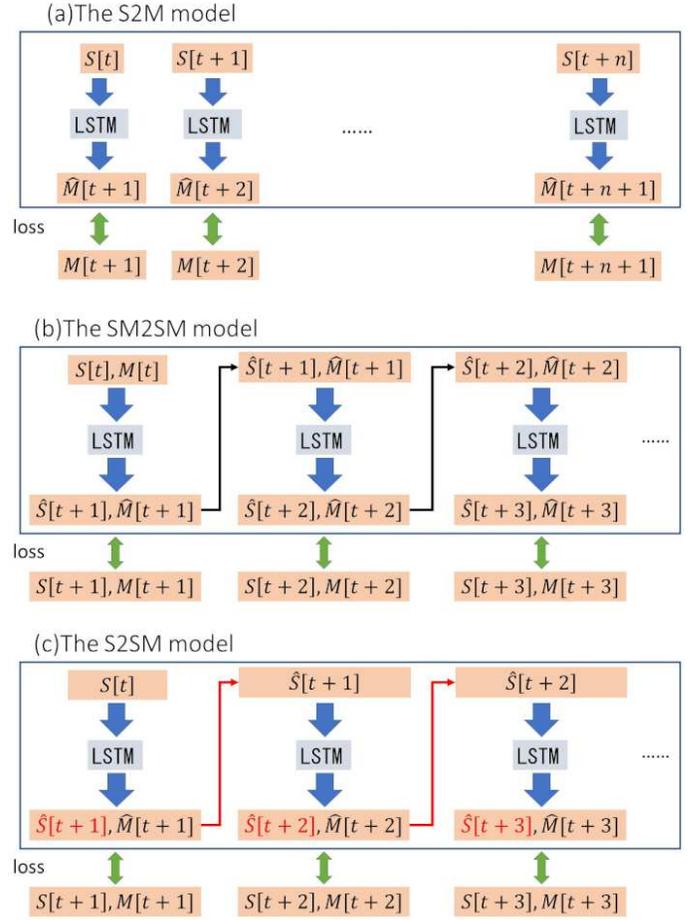}
\caption{Training of models}
\label{fig:auto}
\end{center}
\end{figure}

After the end of data collection, the sampling rate was reduced to 20 ms.
In Fig.~\ref{fig:auto}, $M[t], S[t]$ are the states of the slave and the master, respectively, at each time step.
During the training phase, each time-step was set as $20 \ \mathrm{ms}$. 
Three models compared in this paper are described below.
\subsubsection{S2M Model}
\leavevmode \\
\quad
In the S2M model, which is shown at the top of Fig.~\ref{fig:auto}, the neural network was trained to predict the next state of the master robot from the slave robot's state \cite{b8}.

\subsubsection{SM2SM Model}
\leavevmode \\
\quad When training the SM2SM model, shown in the center of Fig.~\ref{fig:auto}, the output of the model was used as the input of the neural network recursively. This model is good at predicting long-term tasks \cite{auto}.

\subsubsection{S2SM Model (\bf{the proposed method})}
\leavevmode \\
\quad
In the S2SM model, shown at the bottom of Fig.~\ref{fig:auto}, the neural network was trained to predict the next state of the master and slave robots from the slave robot's state.
In this model, by using the predicted state as the input of the model, the far future state can be estimated, which enables the neural networks to plan long-term motions comparable to the SM2SM model.
However, in the SM2SM model, we predict that the appropriate motions cannot be generated because the virtual master state and the measured slave state are used as the input. On the other hand, this problem is solved in the S2SM model because only the slave state is used as the input without using the virtual master state.
In summary, this model is expected to maintain the long-term prediction ability of the SM2SM model and improve stability during autonomous operation.

\subsection{Autonomous Task Execution Phase}
 In this phase, the slave robot executes tasks autonomously. The slave robot measures the response values $S^{res}$ and uses it as inputs of the trained neural networks, as shown in Fig.~\ref{fig:resmodel}. 
 Subsequently, the master outputs of the neural network are used as command values. The neural network repeated this procedure every 20 ms, and the control cycle was 1 ms.

\subsubsection{conventional control method}
\leavevmode \\
\quad
In the conventional method, $\hat M$ of the model is used as the command value of the slave robot as follows:
\begin{eqnarray}
\theta_s^{cmd} &=& \hat\theta_m,\label{eq:conv1} \\
\dot\theta_s^{cmd} &=& {\hat{\dot{\theta}}}_m,\label{eq:conv2} \\
\tau_s^{cmd} &=& \hat\tau_m. \label{eq:conv3}
\end{eqnarray}

\subsubsection{Command Feedback in Autoregressive Models (\bf{the proposed method})}
\leavevmode \\
\quad
Because the true values of the master's behavior cannot be measured in autonomous operation, we cannot calculate the estimation errors in the S2M model. On the contrary, by using autoregressive models such as SM2SM and S2SM, we can obtain both true and estimated values of the slave states, and thus can calculate the estimation errors. Notably, in bilateral control-based imitation learning, the master and slave estimates are strongly related because they are naturally controlled to satisfy the following control goals (1) and (2).
Therefore, in this study, we assumed that the estimation error of the slave is equal to the estimation error of the master, as follows:
\begin{eqnarray}
\theta_m^{res} - \hat\theta_m &=& \theta_s^{res} - \hat\theta_s,\label{eq:relation1} \\ 
\dot\theta_m^{res} - {\hat{\dot{\theta}}}_m &=& \dot\theta_s^{res} - {\hat{\dot{\theta}}}_s,\label{eq:relation2} \\ 
\tau_m^{res} - \hat\tau_m &=& -(\tau_s^{res} - \hat\tau_s). \label{eq:relation3} 
\end{eqnarray}
Moreover, these equations can be transformed as following:
\begin{eqnarray}
\theta_m^{res} &=& \hat\theta_m + (\hat\theta_s - \theta_s^{res}),\label{eq:trre1} \\ 
\dot\theta_m^{res} &=& {\hat{\dot{\theta}}}_m + ({\hat{\dot{\theta}}}_s - \dot\theta_s^{res}),\label{eq:trre2} \\ 
\tau_m^{res} &=& \hat\tau_m - (\hat\tau_s - \tau_s^{res}). \label{eq:trre3} 
\end{eqnarray}
These are interpreted as equations that add the slave's estimation error's feedback term to the conventional control method, which allows us to compensate the command values with the estimation errors caused by environmental changes by adopting (9), (10), and (11) as the new command values. In addition, as shown in (\ref{eq:conv1}), (\ref{eq:conv2}), and (\ref{eq:conv3}), the predicted next state of the master was used as the command value in the conventional method. On the other hand, these new command values are regarded as more appropriate because the master's response values were used as the command values for the slave during the bilateral control in the data collection phase. 
In the experiments, a low-pass filter was used to avoid chattering. Therefore, the command values are defined as follows:
\begin{eqnarray}
  \theta_s^{cmd} &=& LPF \{ \hat\theta_m + (\hat\theta_s -\theta_s^{res} ) \},\label{eq:feed1} \\
  \dot\theta_s^{cmd} &=& LPF \{ {\hat{\dot{\theta}}}_m + ( {\hat{\dot{\theta}}}_s -\dot\theta_s^{res}) \},\label{eq:feed2} \\
  \tau_s^{cmd} &=& LPF \{ \hat\tau_m - (\hat\tau_s -\tau_s^{res}) \}. \label{eq:feed3} 
\end{eqnarray}
The low-pass filter described as $LPF$ was defined as follows:
\begin{equation}
  y_t = K y_{t-1} + (1-K) x_t,
\end{equation}
where the coefficient $K$ was 0.5. Here, $x$ is the input of the low-pass filter and $y$ is the output of $LPF$ at each time step $t$.

\section{EXPERIMENT}
\subsection{Data Collection Phase}
\begin{figure}[t]
\centering
\includegraphics[width=3.0cm]{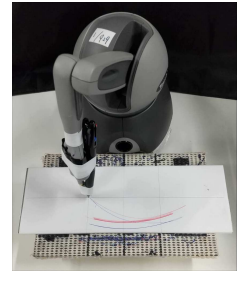}
\caption{Touch$^{TM}$ USB haptic devices used during the autonomous operation}
\label{fig:phantom-autonomous}
\end{figure}

\begin{figure}[t]
\centering
\includegraphics[width=5.0cm]{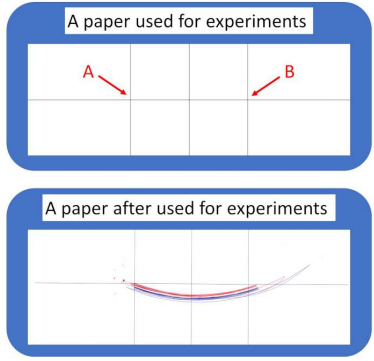}
\caption{A paper for experiments}
\label{fig:paper}
\end{figure}

In this phase, a ballpoint pen was fixed to the slave robot, as shown in Fig.~\ref{fig:phantom-autonomous}, and a human operator teleoperated it through the master robot. Fifteen trials were performed in total, with each trial of 13 s duration. In addition, the heights of the paper samples during collection of data were set at 70 mm, 45 mm, and 19 mm, with five data points collected for each height measurement. 
In this study, the behavioral data required to draw arcs through points A and B were collected as shown in Fig.~\ref{fig:paper}.

\begin{table*}[t]
\centering
\caption{Training Time and Epoch Number of Each Method}
\begin{tabular}{|l|l|l|l|}
\hline
Model                  & Autoregression number & Epoch & Time           \\ \hline
S2M                    & -                     & 1000  & 6min34seconds  \\ \hline
\multirow{3}{*}{SM2SM} & 1                     & 1000  & 6min45seconds  \\ \cline{2-4} 
                        & 5                     & 3000  & 19min53seconds \\ \cline{2-4} 
                        & 10                    & 4000  & 26min30seconds \\ \hline
\multirow{3}{*}{S2SM}  & 1                     & 1000  & 7min1seconds   \\ \cline{2-4} 
                        & 5                     & 3000  & 20min9seconds  \\ \cline{2-4} 
                        & 10                    & 4000  & 26min25seconds \\ \hline
\end{tabular}
\label{tab:traintime}
\end{table*}

\begin{table*}[t]
\centering
\caption{Results of Autonomous Operation at Each Paper Height and Method}
\begin{tabular}{|l|l|l|l|l|l|l|l|l|l|l|l|l|l|}
\hline
\multirow{4}{*}{\begin{tabular}[c]{@{}l@{}}Heights\\  of \\ papers\\ {[}mm{]}\end{tabular}} & \multirow{4}{*}{S2M} & \multicolumn{6}{l|}{S2SM}                                                 & \multicolumn{6}{l|}{SM2SM}                                                \\ \cline{3-14} 
                                                                                            &                      & \multicolumn{6}{l|}{Autoregression number}                                & \multicolumn{6}{l|}{Autoregression number}                                \\ \cline{3-14} 
                                                                                            &                      & \multicolumn{2}{l|}{1} & \multicolumn{2}{l|}{5} & \multicolumn{2}{l|}{10} & \multicolumn{2}{l|}{1} & \multicolumn{2}{l|}{5} & \multicolumn{2}{l|}{10} \\ \cline{3-14} 
                                                                                            &                      & conv        & fb        & conv        & fb        & conv        & fb         & conv        & fb        & conv       & fb         & conv         & fb        \\ \hline
70                                                                                          & \checkmark                  & \checkmark        & \checkmark       & \checkmark        & \checkmark       & \checkmark        & \checkmark        & \checkmark        & \checkmark       & \checkmark       & \checkmark        & -        & -      \\ \hline
55                                                                                          & \checkmark                  & \checkmark        & \checkmark       & \checkmark        & \checkmark       & \checkmark        & \checkmark        & \checkmark        & \checkmark       & \checkmark       & \checkmark        & -        & -      \\ \hline
45                                                                                          & \checkmark                  & \checkmark        & \checkmark       & \checkmark        & \checkmark       & \checkmark        & \checkmark        & \checkmark        & \checkmark       & \checkmark       & \checkmark        & \checkmark         & -      \\ \hline
31                                                                                          & \checkmark                  & \checkmark        & \checkmark       & \checkmark        & \checkmark       & \checkmark        & \checkmark        & \checkmark        & \checkmark       & \checkmark       & -       & \checkmark         & \checkmark       \\ \hline
19                                                                                          & \checkmark                  & \checkmark        & \checkmark       & \checkmark        & \checkmark       & \checkmark        & \checkmark        & -       & \checkmark       & \checkmark       & -       & \checkmark         & \checkmark       \\ \hline
\end{tabular}
\label{tab:allresult}
\end{table*}

\begin{figure}[t]
\centering
\includegraphics[width=7.0cm]{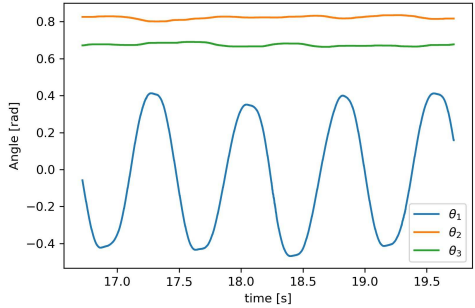}
\caption{Angle responses while drawing arcs}
\label{fig:tau1}
\end{figure}

\subsection{Training Phase}
The above-mentioned three model types (S2M, SM2SM, and S2SM) were executed in the training phase. 
The models were trained on a machine with Ubuntu 18.04 OS, AMD Ryzen 7 3700x 8-core processor, and GeForce RTX 2080. The training time and epoch number of each method are shown in Table \ref{tab:traintime}.

\subsection{Autonomous Task Execution Phase}
The operations were executed autonomously in this phase. The performance of each model was tested at five paper heights (70, 55, 45, 31, and 19 mm). The behavior with command feedback (denoted as fb) was compared with that obtained with the conventional method (denoted as conv). Table~\ref{tab:allresult} shows the success (\checkmark) or failure (-) of the autonomous execution.
As shown in Table~\ref{tab:allresult}, the S2SM model showed a higher success rate compared with the SM2SM model. Here, each number (1, 5, and 10) indicates the number of times the autoregression was repeated in the training phase, and the details of the autoregression are described in Section III-B. As shown in Table~\ref{tab:allresult}, the success rates of the S2M and S2SM models were 100\%, respectively, while that of the SM2SM model was 76.7\%, which means that the use of master states in the inputs of the neural network resulted in poorer stability. It is known that S2M is effective for performing periodic tasks \cite{auto}; the S2SM model also proved to be effective in executing periodic tasks. In addition, S2SM showed superior results to the S2M model in terms of the reproducibility of the training data discussed in Section IV-E.

\subsection{Evaluation of the variance ratio between two methods}
Command feedback is considered to work to compensate for the difference between $\hat S$ and $S^{res}$. Therefore, the variance between $\hat S$ and $S^{res}$ at $\theta_1$, $\theta_2$, $\theta_3$, $\dot\theta_1$, $\dot\theta_2$, $\dot\theta_3$, $\tau_1$, $\tau_2$, and $\tau_3$ were calculated as follows:

\begin{eqnarray}
V_\theta = \frac{1}{T} \sum_{t=1}^{T} (\theta_t^{res} - \hat\theta_t)^2 , \\
V_{\dot\theta} = \frac{1}{T} \sum_{t=1}^{T} (\dot\theta_t^{res} - {\hat{\dot{\theta}}}_t)^2 , \\
V_{\tau} = \frac{1}{T} \sum_{t=1}^{T} (\tau_t^{res} - \hat\tau_t)^2 ,
\end{eqnarray}
where $T$ is the time step when the autonomous operation ended. 
Subsequently, the variance ratio (conventional/feedback) was considered as follows: 
\begin{eqnarray}
    V_{\theta}^{ratio} = \frac{1}{3} \sum_{i=1}^{i=3} \frac{V_{\theta_i}^{conv}}{V_{\theta_i}^{fb}} , \\
    V_{\dot\theta}^{ratio} = \frac{1}{3} \sum_{i=1}^{i=3}  \frac{V_{\dot\theta_i}^{conv} }{ V_{\dot\theta_i}^{fb} } ,\\
    V_{\tau}^{ratio} = \frac{1}{3} \sum_{i=1}^{i=3}  \frac{ V_{\tau_i}^{conv} }{V_{\tau_i}^{fb}} . 
\label{eq:v_1}
\end{eqnarray}
These ratios were higher than 1.0 when the command feedback reduced the differences between $\hat S$ and $S^{res}$. 
For a comprehensive evaluation of each method, the total variance $V^{total}_{ratio}$ was defined as follows: 
\begin{equation}
    V^{total}_{ratio} =  V_{\theta}^{ratio} + V_{\dot\theta}^{ratio} + V_{\tau}^{ratio} . 
\label{eq:v_2}
\end{equation}
Although the evaluation considering this total ratio may not be absolutely correct, it is possible to assess the overall effect of the feedback.
First, the features of the variance ratios of angle, angular velocity, torque, and total variance ratio were described. As shown in Table~\ref{tab:var}, when the autoregression number was 1 or 10,  the effect of the feedback was confirmed in most situations. On the other hand, when the autoregression number was 5, the effect of the feedback was restricted. 
Second, as shown in the result in Section IV-E, the best reproducibility of the motion was observed when the autoregression number was 5. Thus, it is considered that 5 was the appropriate number of autoregressions in this task. 
In other words, when the autoregression number was 5, the S2SM model succeeded in generating desirable motion without the proposed feedback, and
apparently the error $(S^{res}- \hat S)$ also did not decrease.

\begin{figure*}[t]
\centering
\includegraphics[width=16.0cm]{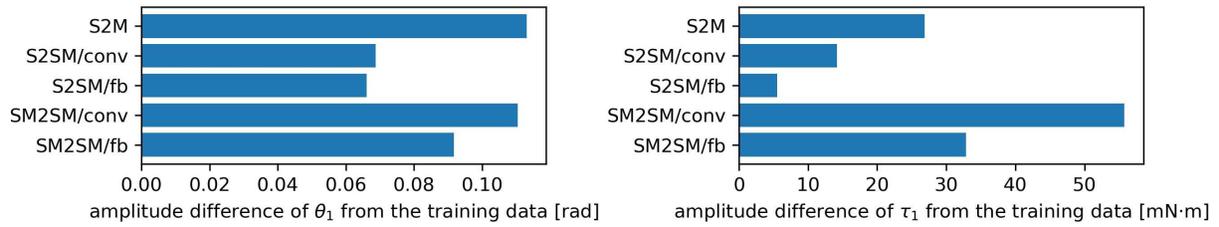}
\caption{The amplitude difference in $\theta_1$ and $\tau_1$ between autonomous operation data and training data at 45 mm paper height}
\label{fig:integ}
\end{figure*}

\begin{table*}[t]
\centering
\caption{variance ratio at each autoregression number and paper height in using the S2SM model}
\begin{tabular}{|l|l|l|l|l|l|l|l|l|l|l|l|l|}
\hline
                                                                                            & \multicolumn{12}{c|}{Autoregression number}                                                                                                                                                                                                                         \\ \cline{2-13} 
                                                                                            & \multicolumn{4}{c|}{1}                    & \multicolumn{4}{c|}{5}                                                                                                & \multicolumn{4}{c|}{10}                                                                         \\ \cline{2-13} 
\multirow{-3}{*}{\begin{tabular}[c]{@{}l@{}}Heights\\  of papers\\  {[}mm{]}\end{tabular}} & $V_{\theta}^{ratio}$ & $V_{\dot\theta}^{ratio}$ & $V_{\tau}^{ratio}$ & $V_{ratio}^{total}$ & $V_{\theta}^{ratio}$                       & $V_{\dot\theta}^{ratio}$            & $V_{\tau}^{ratio}$                      & $V_{ratio}^{total}$                       & $V_{\theta}^{ratio}$                       & $V_{\dot\theta}^{ratio}$            & $V_{\tau}^{ratio}$                      & $V_{ratio}^{total}$ \\ \hline
70                                                                                         & 2.03  & 1.20             & 2.08   & \bf{1.77}  & 1.46 &  0.71 &  0.83 & \bf{1.00} & 1.49 & 1.12 & 1.51 & \bf{1.37}  \\ \hline
55                                                                                         & 10.86 & 1.32             & 7.53   & \bf{6.57}  & 1.11 &  0.71 &  0.65 & 0.82 & 1.04  & 1.07   & 1.00    & \bf{1.04}  \\ \hline
45                                                                                         & 2.05  & 1.07             & 3.32   & \bf{2.15}  & 0.93 &  0.70 &  0.86 & 0.83 & 1.09  & 1.09 &  0.98 & \bf{1.05}  \\ \hline
31                                                                                         & 1.40  & 1.07             & 2.94   & \bf{1.80}  & 1.77 &  0.80 & 1.32  & \bf{1.30} &  0.95 & 1.31                        &  0.98 & \bf{1.08}  \\ \hline
19                                                                                         & 2.91  & 1.99             & 6.03   & \bf{3.64}  & 1.36 & 1.06  & 1.33  & \bf{1.25} &  0.93 &  0.95 & 2.41                        & \bf{1.43}  \\ \hline
\end{tabular}
\label{tab:var}
\end{table*}

\subsection{Comparison between training data and the result of autonomous operations}
To assess the reproducibility of the training models and the proposed feedback method, it is necessary to compare the training data and the results of autonomous operations.
In performing this task, the movement of the first axis ($\theta_1$) was periodical just like a sine wave. The second and third axes ($\theta_2$ and $\theta_3$) were more constant than the first, as shown in Fig.~ \ref{fig:tau1}.
Therefore, the amplitude and average of $\theta_1$ and $\tau_1$ were calculated from the training data and autonomous operation data at a paper height of 45 mm.
Then, absolute differences between the training and autonomous operation data were compared. 
In this experiment, the S2SM and SM2SM models were tested for autoregression numbers of 1, 5, and 10. 
As a result, the absolute differences in the amplitudes of $\theta_1$ and $\tau_1$ were minimum when the autoregression number was 5. Therefore, amplitude differences in the case of five autoregressions  are shown in Fig~\ref{fig:integ}.
To obtain the absolute difference in the amplitudes of $\theta_1$ shown on the left side of Fig.~\ref{fig:integ}, the amplitude difference was reduced using the S2SM model. Moreover, to observe the absolute difference in the amplitudes of $\tau_1$ shown on the right side of Fig.~\ref{fig:integ}, the amplitude difference was reduced with the command feedback. 
Notably, in the S2SM model, estimation errors did not accumulate because the inputs were measured responses without any estimates. Simultaneously, the command feedback also improved the reproducibility because the estimation errors of the master states were approximated and measured by the estimated errors of the slave states.

\section{CONCLUSION}
In this study, the characteristics of three learning methods, S2M, S2SM, and SM2SM, and a control method with and without feedback were examined. 
The S2SM model displayed a higher probability of success in executing tasks than the SM2SM model. In addition, the S2SM model was effective in not only reducing errors between a neural network's predictive value $\hat S$ and the secondary robot's response $S^{res}$ but also reproducing training data more accurately than other methods. In summary, the S2SM model demonstrated better results than the conventional methods, S2M and SM2SM.

\section*{Acknowledgment}
This work was supported by JST PRESTO Grant Number JPMJPR1755, Japan. 
This research is also supported by Adaptable and Seamless Technology transfer Program through Target-driven R\&D (A-STEP) from Japan Science and Technology Agency(JST).

\vspace{12pt}
\color{red}

\begin{thebibliography}{00}
\bibitem{b2} T. Tsuji, K. Kutsuzawa, and S. Sakaino ``Optimized trajectory generation based on model predictive control for turning over pancakes,” {\it IEEJ Journal of Industry Applications}, vol.~7, no.~1, pp.~22--28, 2018.
\bibitem{b3} K. Tsuda, T. Sakuma, K. Umeda, S. Sakaino, and T. Tsuji ``Resonance-suppression control for electro-hydrostatic actuator as two-inertia system,” {\it IEEJ Journal of Industry Applications}, vol.~6, no.~5, pp.~320-327, 2017.
\bibitem{levine} S. Levine, P. Pastor, A. Krizhevsky, J. Ibarz, and D. Quillen, ``Learning handEye coordination for robotic grasping with deep learning and large-scale data collection," {\it The International Journal of Robotics
Research}, vol.~37, no.~4–-5, pp.~421--436, 2018.
\bibitem{sim2real}J. Tobin, R. Fong, A. Ray, J. Schneider, W. Zaremba, and P. Abbeel, ``Domain randomization for transferring deep neural networks from simulation to the real world," in {\it Proceedings of the IEEE/RSJ International Conference on Intelligent Robots and Systems}, pp.~23--30, 2017.
\bibitem{offline} S. Levine, A. Kumar, G. Tucker, and J. Fu, ``Offline reinforcement learning: Tutorial, review, and perspectives on open problems", {\it arXiv:2005.01643}, 2020.
\bibitem{survey} B. Fang, S. Jia, D. Guo, M. Xu, S. Wen, and F. Sun, ``Survey of imitation learning for robotic manipulation," {\it 2019 International Journal of Intelligent Robotics and Applications}, vol.~3, no.~4, pp.~362--369, 2019.
\bibitem{survey2} A. Hussein, M. Gaber, E. Elyan, and C. Jayne, ``Imitation learning: A survey of learning methods," {\it ACM Computing Surveys}, vol.~50, no.~2, pp.~35, 2017.
\bibitem{motioncap} A. P. Shon, K. Grochow, and R. P. N. Rao, ``Robotic imitation from human motion capture using gaussian processes," in {\it Proceedings of the 5th IEEE-RAS International Conference on Humanoid Robots}, pp.~129--134, 2005.
\bibitem{vr} T. Zhang, Z. McCarthy, O. Jow, D. Lee, X. Chen, K. Goldberg, and P. Abbeel, ``Deep imitation learning for complex manipulation tasks from virtual reality teleoperation," in {\it Proceedings of 2018 IEEE International Conference on Robotics and Automation}, pp.~5628--5635, 2018.
\bibitem{vr2} J. S. Dyrstad, E. R. Oye, A. Stahl, and J. R. Mathiassen, ``Teaching a robot to grasp real fish by imitation learning from a human supervisor in virtual reality," in {\it Proceedings of 2018 IEEE/RSJ International Conference on Intelligent Robots and Systems}, pp.~7185--7192, 2018.
\bibitem{f1} P. Kormushev, S. Calinon, and D. G. Caldwell, ``Imitation learning of positional and force skills demonstrated via kinesthetic teaching and haptic input," {\it Advanced Robotics}, vol.~25, no.~5, pp.~581--603, 2011.
\bibitem{f2} A. X. Lee, H. Lu, A. Gupta, S. Levine, and P. Abbeel, ``Learning force-based manipulation of deformable objects from multiple demonstrations," in {\it Proceedings of 2015 IEEE International Conference on Robotics and Automation}, pp.~177--184, 2015.
\bibitem{f3} M. Edmonds, F. Gao, X. Xie, H. Liu, S. Qi, Y. Zhu, B. Rothrock, and S. Zhu, ``Feeling the force: Integrating force and pose for fluent discovery through imitation learning to open medicine bottles," in {\it Proceedings of 2017 IEEE/RSJ International Conference on Intelligent Robots and Systems}, pp.~3530--3537, 2017.
\bibitem{b9} K. Fujimoto, S. Sakaino, and T. Tsuji, ``Time series motion generation considering long short-term motion," in {\it Proceedings of 2019 IEEE/RSJ International Conference on Intelligent Robots and Systems}, pp.~6842--6848, 2019.
\bibitem{b10} T. Adachi, K. Fujimoto, S. Sakaino, and T. Tsuji, ``Imitation learning for object manipulation based on position/force information using bilateral control," in {\it Proceedings of 2018 IEEE/RSJ International Conference on Intelligent Robots and Systems}, pp.~3648--3653, 2019.
\bibitem{b8} A. Sasagawa, K. Fujimoto, S. Sakaino, and T. Tsuji, ``Imitation learning based on bilateral control for human–robot cooperation," {\it IEEE Robotics and Automation Letters}, vol.~5, no.~4, pp.~6169--6176, 2020.
\bibitem{bilate1} T. Kitamura, S. Sakaino, and T. Tsuji, ``Bilateral control using functional electrical stimulation," in {\it Proceedings of the 41st Annual Conference of the IEEE Industrial Electronics Society}, pp.~2336--2341, 2015.
\bibitem{bilate2} S. Sakaino, T. Furuya, and T. Tsuji, ``Bilateral control between electric and hydraulic actuators using linearization of hydraulic actuators," {\it IEEE Transactions on Industrial Electronics}, vol.~64, no.~6, pp.~4631-4641, 2017.
\bibitem{bilate3} S. Sakaino, T. Sato, and K. Ohnishi, ``Oblique coordinate control for advanced motion control-applied to micro-macro bilateral control," in {\it Proceedings of IEEE International Conference on Mechatronics}, pp.~1--6, 2009.
\bibitem{b1} T. Kitamura, N. Mizukami, S. Sakaino, and T. Tsuji, ``Bilateral control in the vertical direction using functional electrical stimulation,” {\it IEEJ Journal of Industry Applications}, vol.~5, no.~5, pp.~398--404, 2016.
\bibitem{auto} A. Sasagawa, S. Sakaino, and T. Tsuji, ``Motion generation using bilateral control-based imitation learning with autoregressive learning," {\it arXiv:2011.06192}, 2020.
\bibitem{dob} K. Ohnishi, M. Shibata, and T. Murakami, ``Motion control for advanced mechatronics," {\it IEEE/ASME Transaction on Mechatronics}, vol.~1, no.~1, pp.~56--67, 1996.
\bibitem{rfob} T. Murakami, F. Yu, and K. Ohnishi, ``Torque sensorless control in multidegree-of-freedom manipulator," {\it IEEE Transactions on Industrial Electronics}, vol.~40, no.~2, pp.~259--265, 1993.
\bibitem{estim} T. Yamazaki, S. Sakaino, and  T. Tsuji, ``Estimation and kinetic modeling of human arm using wearable robot arm," {\it Electr Eng Jpn}, vol.~199, no.~3, pp.~57--67, 2017.
\end{thebibliography}
\end{document}